\documentclass[10pt,twocolumn,letterpaper]{article}

\usepackage{cvpr}
\usepackage{times}
\usepackage{epsfig}
\usepackage{graphicx}
\usepackage{amsmath}
\usepackage{amssymb}

\usepackage{booktabs}       
\usepackage{amsfonts}       
\usepackage{nicefrac}       
\usepackage{microtype}

\usepackage{graphicx}
\usepackage{multirow}
\usepackage{array}
\usepackage{rotating}
\usepackage{amssymb}
\usepackage{amsmath}
\usepackage{wrapfig}
\usepackage{tabularx}
\usepackage{pifont}

\usepackage[pagebackref=true,breaklinks=true,letterpaper=true,colorlinks,bookmarks=false]{hyperref}

 \cvprfinalcopy 

\def\cvprPaperID{2882} 

\ifcvprfinal\fi
\begin{document}

	\title{Context-Aware Domain Adaptation in Semantic Segmentation}
	\author{Jinyu Yang\textsuperscript{1}, Weizhi An\textsuperscript{1}, Chaochao Yan\textsuperscript{1}, Peilin Zhao\textsuperscript{2}, Junzhou Huang\textsuperscript{1}\\
		\textsuperscript{1}The University of Texas at Arlington\\
		\textsuperscript{2}Tencent AI Lab\\
	}

	\maketitle

	\begin{abstract}
		In this paper, we consider the problem of unsupervised domain adaptation in the semantic segmentation. There are two primary issues in this field, \ie, what and how to transfer domain knowledge across two domains. Existing methods mainly focus on adapting domain-invariant features (what to transfer) through adversarial learning (how to transfer). Context dependency is essential for semantic segmentation, however, its transferability is still not well understood. Furthermore, how to transfer contextual information across two domains remains unexplored. Motivated by this, we propose a cross-attention mechanism based on self-attention to capture context dependencies between two domains and adapt transferable context. To achieve this goal, we design two cross-domain attention modules to adapt context dependencies from both spatial and channel views. Specifically, the spatial attention module captures local feature dependencies between each position in the source and target image. The channel attention module models semantic dependencies between each pair of cross-domain channel maps. To adapt context dependencies, we further selectively aggregate the context information from two domains. The superiority of our method over existing state-of-the-art methods is empirically proved on "GTA5 to Cityscapes" and "SYNTHIA to Cityscapes".

	\end{abstract}

	\section{Introduction}

	Semantic segmentation aims to predict pixel-level labels for the given images \cite{long2015fully, chen2018deeplab}, which has been widely recognized as one of the fundamental tasks in computer vision. Unfortunately, the manual pixel-wise annotation for large-scale segmentation datasets is extremely time-consuming and requires massive amounts of labor efforts. As a tradeoff, synthetic datasets \cite{richter2016playing, ros2016synthia} with freely-available labels offer a promising alternative by providing considerable data for model training. However, the domain discrepancy between synthetic (source) and real (target) images is still the central challenge to effectively transfer knowledge across domains. To overcome this limitation, the key idea of existing methods is to leverage knowledge from a source domain to enhance the learning performance of a target domain. Such a strategy is mainly inspired by the recent advances in unsupervised domain adaptation for image classification \cite{pan2009survey}.


	\begin{figure}[t]
		\begin{center}
			\includegraphics[width=0.95\linewidth]{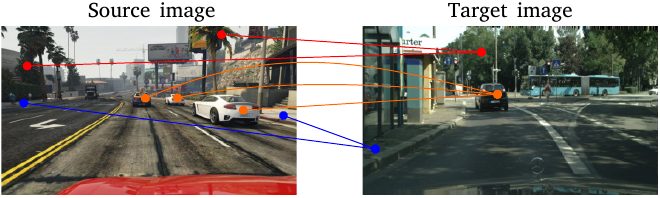}
		\end{center}
		\caption{An example of cross-domain context. The source and target images share similar context information at the spatial and semantic level. The red line, orange line, and blue line denote vegetation, car, and sidewalk across two domains, respectively.}
		\label{fig:motivation}
		\vspace{-0.2in}
	\end{figure}

	Conventional domain adaptation methods in image classification attempt to learn domain-invariant feature representations by directly minimizing the representation distance between two domains \cite{tzeng2014deep, long2015learning, long2017deep}, encouraging a common feature space through an adversarial objective \cite{ganin2014unsupervised, tzeng2017adversarial}, or automatically determining what and where to transfer via meta-learning \cite{ying2018transfer, jang2019learning}. Motivated by this, various domain adaptation methods for semantic segmentation are proposed recently. Among them, the most common practices are based on feature alignment \cite{hoffman2016fcns}, structured output adaptation \cite{tsai2018learning}, curriculum adaptation \cite{zhang2017curriculum, lian2019constructing}, self training \cite{zou2018unsupervised, li2019bidirectional}, and image-to-image translation \cite{hoffman2017cycada, li2019bidirectional, chen2019crdoco, chen2019learning}. Despite remarkable performance improvement achieved by these methods, they fail to explicitly consider the contextual dependencies across the source and target domains which is essential for scene understanding \cite{zhang2018context, zhao2018psanet}. As illustrated in Figure~\ref{fig:motivation}, the source and target images share a much similar semantic context such as vegetation, car, and sidewalk, although their appearances (\eg, scale, texture, and illumination) are quite different. However, how to adapt context information across two domains remains unexplored.

	\begin{figure*}
		\begin{center}
			\includegraphics[width=1.0\linewidth]{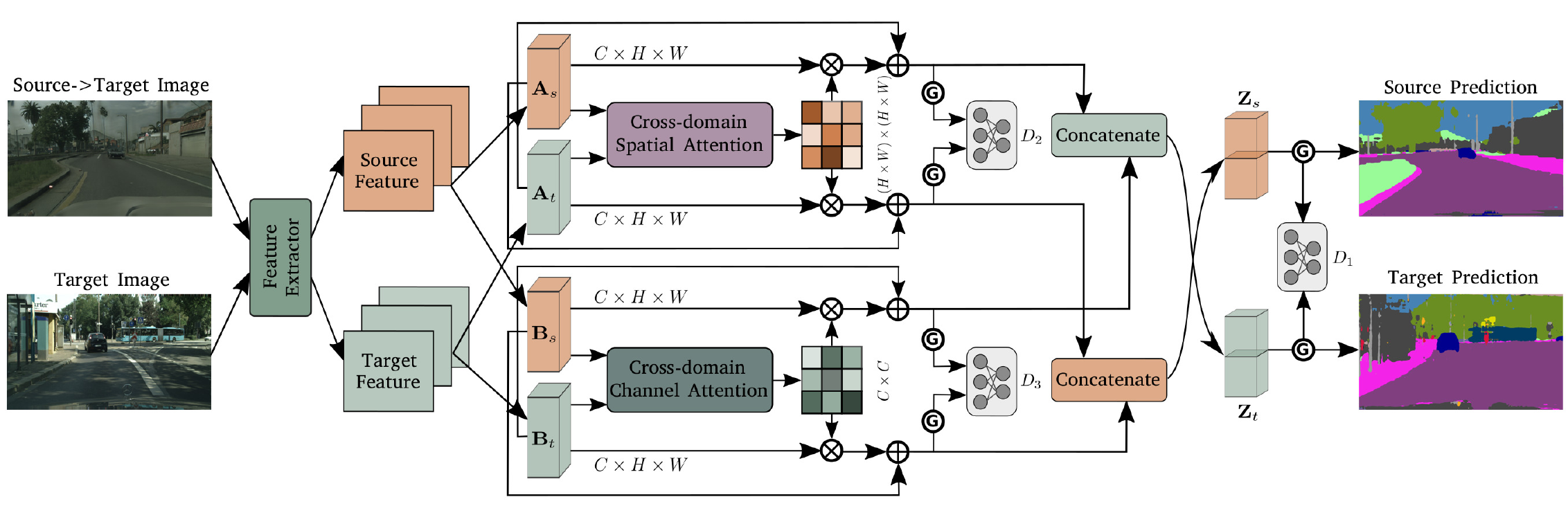}
		\end{center}
		\caption{An overview of the proposed framework. It applies a feature extractor (\ie, ResNet101 or VGG16) to learn source and target features. Two cross-domain attention modules (\ie, CD-SAM and CD-CAM) are designed to adapt spatial and semantic context information across source and target domains. A classifier $ G $ is used to predict segmentation output based on the features from CD-SAM and CD-CAM. Our framework contains three discriminators (\ie, $ D_1 $, $ D_2 $, and $ D_3 $) for output adaptation by enforcing the source output be indistinguishable from the target output.}
		\label{fig:framework}
		\vspace{-0.2in}
	\end{figure*}

	Inspired by this, we propose a novel domain adaptation framework named cross-domain attention network (CDANet), designed for urban-scene semantic segmentation. The key idea of CDANet is to leverage cross-domain context dependencies from both a local and global perspective. To achieve this goal, we innovatively design a cross-attention mechanism which contains two cross-domain attention modules to capture mutual context dependencies between source and target domains. Given that same objects with different appearances and scales often share similar features, we introduce a cross-domain spatial attention module (CD-SAM) to capture local feature dependencies between any two positions in a source image and a target image. The CD-SAM involves two directions (\ie, "source-to-target" and "target-to-source") to adaptively aggregate cross-domain features to learn common context information. On the forward direction (or "source-to-target"), CD-SAM updates the feature at each position in the source image as the weighted sum of features at all positions in the target image. The weights are computed based on the similarity of source and target features at each position. Similarly, the backward direction (or "target-to-source") updates the target feature at each position based on the attention to features at all positions in the source image. In consequence, spatial contexts from the source domain are encoded in the target domain, and vice versa. To model the associations between different semantic responses across two domains, we introduce a cross-domain channel attention module (CD-CAM) which has the same bidirectional structure as CD-SAM. The CD-CAM is designed for contextual information aggregation through capturing the channel feature dependencies between any two channel maps in the source and target image. In such a way, common semantic contexts are shared by both domains. CD-SAM and CD-CAM play a complementary role for context adaptation and their outputs are further merged to provide better feature representations for scene understanding.

	Our main contributions are summarized as follows:
	\begin{itemize}
		\item We propose a novel cross-attention mechanism that enables to transfer of context dependencies across two domains. This is the first-of-its-kind study that investigates the transferability of context information in the domain adaptation.
		\item Two cross-domain attention modules are proposed to capture and adapt context dependencies at both spatial and channel levels. This allows us to learn the common semantic context shared by source and target domains.
		\item Comprehensive empirical studies demonstrate the superiority of our method over the existing state of the art on two benchmark settings, \ie, "GTA5 to Cityscapes" and "SYNTHIA to Cityscapes".
	\end{itemize}

	\section{Related Work}
	\paragraph{Domain Adaptation for Semantic Segmentation}

	Inspired by the Generative Adversarial Network \cite{goodfellow2014generative}, Hoffman \etal \cite{hoffman2016fcns} propose the first domain adaptation model for semantic segmentation by learning domain-invariant features through adversarial training. To rule out task-independent factors during feature alignment, SIBAN \cite{luo2019significance} purifies significance-aware features before the adversarial adaptation to facilitate feature adaptation and stabilize the adversarial training. However, these global adversarial methods ignore to align the category-level joint distribution, which may disturb well-aligned features. To alleviate this problem, Luo \etal propose a category-level adversarial network to encourage local semantic consistency through reweighting the adversarial loss for each feature \cite{luo2019taking}. Based on the hypothesis that structure information plays an essential role in semantic segmentation, Chang \etal adapt structure information by learning domain-invariant structure \cite{chang2019all}. This is achieved by disentangling the domain-invariant structure of a given image from its domain-specific texture information. AdaptSetNet moves forward by further considering structured output adaptation which is based on the observation that segmentation outputs of the source and target domains share substantial similarities \cite{tsai2018learning}. Different from AdaptSetNet, we apply three domain discriminators to perform output adaptation on the segmentation outputs from CD-SAM, CD-CAM, and the aggregation of these two modules.

	Most recently, image-to-image translation \cite{zhu2017unpaired} has proved its effectiveness in domain adaptation \cite{hoffman2017cycada, wu2018dcan, chen2019crdoco}. The key idea is to translate images from the source domain to the target domain by using an image translation model and use the translated images for adapting cross-domain knowledge through a segmentation adaptation model. Rather than keeping the image translation model unchanged after obtaining translated images, BDL \cite{li2019bidirectional} applies a bidirectional learning framework to alternatively optimize the image translation model and the segmentation model. Similar to \cite{zou2018unsupervised}, a self-supervised learning strategy is also used in BDL to generate pseudo labels for target images and re-training the segmentation model with these labels. Although BDL achieves the new state of the art, it is limited in its ability to consider the cross-domain context dependencies. To overcome this limitation, we introduce two cross-domain attention modules to adapt context information between source and target domains.

	\begin{figure}[t]
		\begin{center}
			\includegraphics[width=0.85\linewidth]{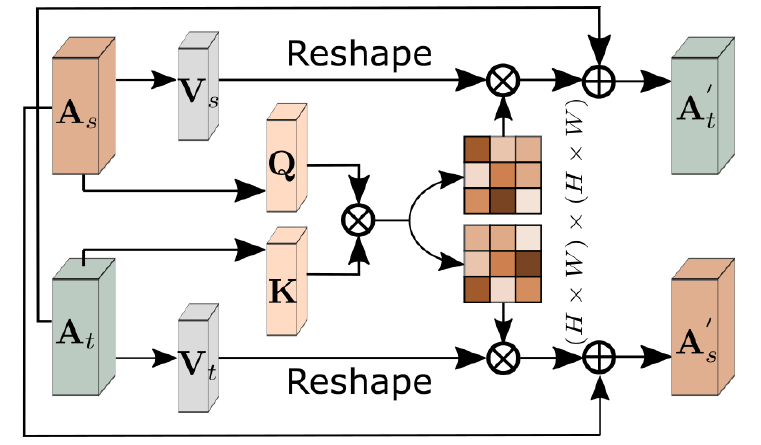}
		\end{center}
		\caption{Cross-domain spatial attention module.}
		\label{fig:spatial_attention}
		\vspace{-0.2in}
	\end{figure}

	\paragraph{Context-Aware Embedding}

	It has been long known that context information plays an important role in perceptual tasks such as semantic segmentation \cite{mottaghi2014role}. Zhang \etal \cite{zhang2018context} propose a context encoding module to capture the semantic context of scenes and selectively emphasize or de-emphasize class-dependent feature maps. To aggregate image-adapted context, MSCI \cite{lin2018multi} further considers multi-scale context embedding and spatial relationships among super-pixels in a given image. Following the success of attention mechanism \cite{vaswani2017attention} in image generation \cite{zhang2018self} and sentence embedding \cite{lin2017structured}, recent studies have highlighted the potential of self-attention in capturing context dependencies \cite{fu2019dual, zhao2018psanet}. Specifically, Zhao \etal \cite{zhao2018psanet} introduce a point-wise spatial attention network to aggregate long-range contextual information. Their model mainly draws its strength from the self-adaptively predicted attention maps which can take full advantage of both nearby and distant information of each pixel. DANet \cite{fu2019dual} adaptively integrates local features with their global dependencies through a position attention module and a channel attention module. These two modules are considered to be able to capture spatial and semantic interdependencies, and in turn, facilitate scene understanding. As opposed to capturing contextual information within a single domain as previously reported, we design an innovative cross-attention mechanism to model context dependencies between two different domains, which is essential for context adaptation.

	\begin{figure}[t]
		\begin{center}
			\includegraphics[width=0.85\linewidth]{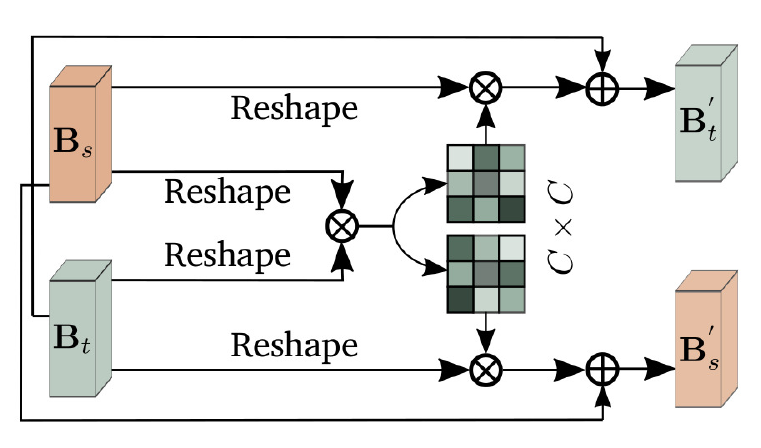}
		\end{center}
		\caption{Cross-domain channel attention module.}
		\label{fig:channel_attention}
		\vspace{-0.2in}
	\end{figure}

	\section{Methodology}

	In this section, we begin by briefing the key idea of our framework. We then detail the proposed cross-attention mechanism which contains two cross-domain attention modules for adapting context dependencies between a source and a target domain.

	\begin{table*}[t]
		\caption{The performance comparison by adapting from GTA5 to Cityscapes. Two base architectures (i.e., VGG16 and ResNet101) are used in our study. The comparison is performed on 19 common classes between source and target domains. We use per-class IoU and mean IoU (mIoU) for the performance measurement. The best result in each column is highlighted in bold.}
		\label{table:gta2city}

		\footnotesize
		\setlength\tabcolsep{3pt}
		\begin{center}
			\begin{tabular}{ @{} l|c|*{19}{c}|*{1}{c} @{} }
				\toprule
				\multicolumn{22}{ c }{\bf GTA5 to Cityscapes } \\
				\midrule
				& \rotatebox[origin=c]{90}{Architecture} & \rotatebox[origin=c]{90}{road} & \rotatebox[origin=c]{90}{sidewalk} & \rotatebox[origin=c]{90}{building} & \rotatebox[origin=c]{90}{wall} & \rotatebox[origin=c]{90}{fence} & \rotatebox[origin=c]{90}{pole} & \rotatebox[origin=c]{90}{  traffic light  } & \rotatebox[origin=c]{90}{traffic sign} & \rotatebox[origin=c]{90}{vegetation} & \rotatebox[origin=c]{90}{terrain} & \rotatebox[origin=c]{90}{sky} & \rotatebox[origin=c]{90}{person} & \rotatebox[origin=c]{90}{rider} & \rotatebox[origin=c]{90}{car} & \rotatebox[origin=c]{90}{truck} & \rotatebox[origin=c]{90}{bus} & \rotatebox[origin=c]{90}{train} & \rotatebox[origin=c]{90}{motorbike} & \rotatebox[origin=c]{90}{bicycle} & \rotatebox[origin=c]{90}{\bf mIoU} \\
				\midrule
				FCNs wild \cite{hoffman2016fcns} & \multirow{8}{*}{\rotatebox[origin=c]{90}{VGG16}} &
				70.4 & 32.4 & 62.1 & 14.9 & 5.4 & 10.9 & 14.2 & 2.7 & 79.2 & 21.3 & 64.6 & 44.1 & 4.2 & 70.4 & 8.0 & 7.3 & 0.0 & 3.5 & 0.0 & 27.1 \\

				CDA \cite{zhang2017curriculum} &  &
				74.9 & 22.0 & 71.4 & 6.0 & 11.9 & 8.4 & 16.3 & 11.1 & 75.7 & 13.3 & 66.5 & 38.0 & 9.3 & 55.2 & 18.8 & 18.9 & 0.0 & 16.8 & 14.6 & 28.9 \\

				AdaptSegNet \cite{tsai2018learning} &  &
				87.3 & 29.8 & 78.6 & 21.1 & 18.2 & 22.5 & 21.5 & 11.0 & 79.7 & 29.6 &
				71.3 & 46.8 & 6.5 & 80.1 & 23.0 & 26.9 & 0.0 & 10.6 & 0.3 & 35.0 \\

				CyCADA \cite{hoffman2017cycada} &  &
				85.2 & 37.2 & 76.5 & 21.8 & 15.0 & \bf 23.8 & 22.9 &  21.5 & 80.5 & 31.3 &
				60.7 & 50.5 & 9.0 & 76.9 & 17.1 &  28.2 & 4.5 & 9.8 & 0.0 & 35.4 \\

				LSD \cite{sankaranarayanan2018learning} &  &
				88.0 & 30.5 & 78.6 & 25.2 & 23.5 & 16.7 & 23.5 & 11.6 & 78.7 & 27.2 & 71.9 & 51.3 & 19.5 & 80.4 & 19.8 & 18.3 & 0.9 & 20.8 & 18.4 & 37.1 \\

				CrDoCo \cite{chen2019crdoco} &  &
				89.1 & 33.2 & 80.1 & 26.9 & 25.0 & 18.3 & 23.4 & 12.8 & 77.0 & 29.1 & 72.4 & 55.1 & 20.2 & 79.9 & 22.3 & 19.5 & 1.0 & 20.1 & 18.7 & 38.1 \\

				BDL \cite{li2019bidirectional} &  &
				89.2 & 40.9 & 81.2 & 29.1 & 19.2 & 14.2 & 29.0 & 19.6 & 83.7 & 35.9 &
				\bf 80.7 & 54.7 & 23.3 & 82.7 & \bf 25.8 & 28.0 & 2.3 & \bf 25.7 & 19.9 & 41.3 \\

				Ours &  &
				\bf 90.1 & \bf 46.7 & \bf 82.7 & \bf 34.2 & \bf 25.3 & 21.3 & \bf 33.0 & \bf 22.0 & \bf 84.4 & \bf 41.4 &
				78.9 & \bf 55.5 & \bf 25.8 & \bf 83.1 & 24.9 & \bf 31.4 & \bf 20.6 & 25.2 & \bf 27.8 & \bf 44.9 \\

				\midrule

				AdaptSegNet \cite{tsai2018learning} & \multirow{5}{*}{\rotatebox[origin=c]{90}{ResNet101}} &
				86.5 & 36.0 & 79.9 & 23.4 & 23.3 & 23.9 & 35.2 & 14.8 & 83.4 & 33.3 &
				75.6 & 58.5 & 27.6 & 73.7 & 32.5 & 35.4 & 3.9 & 30.1 & 28.1 & 42.4 \\

				CLAN \cite{luo2019taking} &  &
				87.0 & 27.1 & 79.6 & 27.3 & 23.3 & 28.3 & 35.5 & 24.2 & 83.6 & 27.4 & 74.2 &
				58.6 & 28.0 & 76.2 & 33.1 & 36.7 & \bf 6.7 & 31.9 & 31.4 & 43.2 \\

				MaxSquare \cite{maxsquareloss} &  &
				89.4 & 43.0 & 82.1 & 30.5 & 21.3 & 30.3 & 34.7 & 24.0 & \bf 85.3 & 39.4 & 78.2 &
				\bf 63.0 & 22.9 & \bf 84.6 & \bf 36.4 & 43.0 & 5.5 & \bf 34.7 & 33.5 & 46.4 \\

				BDL \cite{li2019bidirectional} &  &
				91.0 & 44.7 & 84.2 & \bf 34.6 & 27.6 & 30.2 & \bf 36.0 & 36.0 & 85.0 & \bf 43.6 &
				\bf 83.0 & 58.6 & 31.6 & 83.3 & 35.3 & \bf 49.7 & 3.3 & 28.8 & 35.6 & 48.5 \\

				Ours &  &
				\bf 91.3 & \bf 46.0 & \bf 84.5 & 34.4 & \bf 29.7 & \bf 32.6 & 35.8 & \bf 36.4 & 84.5 & 43.2 &
				\bf 83.0 & 60.0 & \bf 32.2 & 83.2 & 35.0 & 46.7 & 0.0 & 33.7 & \bf 42.2 & \bf 49.2 \\

				\bottomrule
			\end{tabular}
		\end{center}
		\vspace{-0.3in}
	\end{table*}

	\subsection{Overview}

	Given a set of source-domain images $ \mathcal{X}_s $ with pixel-wise labels $ Y_s $ and a set of target-domain images $ \mathcal{X}_t $ without any annotation. Our goal is to train a segmentation model that can provide accurate prediction to $ \mathcal{X}_t $. To achieve this, $ \mathcal{X}_s $ is first translated from the source domain to the target domain using CycleGAN \cite{zhu2017unpaired}. The translated images $ \mathcal{X}^{'}_s=\mathcal{F}(\mathcal{X}_s) $ (where $ \mathcal{F} $ denotes the image translation model) share the same semantic labels with $ \mathcal{X}_s $ but with common visual appearance as $ \mathcal{X}_t $. Motivated by the self-training strategy, we follow the same idea in \cite{li2019bidirectional} to generate pseudo labels $ Y^{st}_t $ for $ \mathcal{X}_t $ with high prediction confidence. Coordinated with these translated images and pseudo labels, we introduce a cross-attention mechanism for domain adaptation of semantic segmentation by leveraging cross-domain contextual information (Figure~\ref{fig:framework}). First, a feature extractor $ E $ is applied to get source feature $ E(\mathcal{X}^{'}_s) $ and target feature $ E(\mathcal{X}_t) $ which are 1/8 of the corresponding input image size. Then a linear interpolation is applied to $ E(\mathcal{X}^{'}_s) $ and $ E(\mathcal{X}_t) $ to match their spatial size. After that, two parallel convolution layers are applied to $ E(\mathcal{X}^{'}_s) $ and $ E(\mathcal{X}_s) $ to generate feature pairs $ \{\textbf{A}_{s}, \textbf{A}_{t}\} $ and $ \{\textbf{B}_{s}, \textbf{B}_{t}\} $, respectively. $ \{\textbf{A}_{s}, \textbf{A}_{t}\} $ is then fed into CD-SAM to adapt spatial-level context, while CD-CAM adapts channel-level context based on $ \{ \textbf{B}_{s}, \textbf{B}_{t}\} $.

	For each module, two directions, \ie, forward direction ("source-to-target") and backward direction ("target-to-source") are involved. Take the CD-SAM as an example, an energy map is first obtained based on $ \{\textbf{A}_{s}, \textbf{A}_{t}\} $. This energy map is further divided into two attention matrices denoted by $ \Gamma_{s{\rightarrow}t} $ and $ \Gamma_{t{\rightarrow}s} $. During the forward direction, we perform a matrix multiplication between target features and $ \Gamma_{s{\rightarrow}t} $. The result is then summed with the original source features in an element-wise manner. For the backward direction, a matrix multiplication is conducted between source features and $ \Gamma_{t{\rightarrow}s} $. After that, an element-wise summation between the obtained results and original target features is carried out. The CD-CAM follows the same setting above except that the energy map is calculated in the channel dimension. The final source feature and target feature are obtained by aggregating the outputs from these two attention modules, which are then fed into a classifier $ G $ for semantic segmentation.


	\newcommand{\xmark}{\ding{55}}%
	\begin{table*}[t]
		\caption{The performance comparison by adapting from SYNTHIA to Cityscapes. Two base architectures (i.e., VGG16 and ResNet101) are used in our study. The comparison is performed on 16 common classes for VGG16 and 13 common classes for ResNet101. We use per-class IoU and mean IoU (mIoU) for the performance measurement. The best result in each column is highlighted in bold.}
		\label{table:synthia2city}

		\footnotesize
		\setlength\tabcolsep{2.3pt}
		\begin{center}
			\begin{tabular}{ @{} l|c|*{16}{c}|*{1}{c} @{} }
				\toprule
				\multicolumn{19}{ c }{\bf SYNTHIA to Cityscapes } \\
				\midrule\

				& \rotatebox[origin=c]{90}{Architecture} & \rotatebox[origin=c]{90}{road} & \rotatebox[origin=c]{90}{sidewalk} & \rotatebox[origin=c]{90}{building} & \rotatebox[origin=c]{90}{wall} & \rotatebox[origin=c]{90}{fence} &\rotatebox[origin=c]{90}{pole} & \rotatebox[origin=c]{90}{traffic light} & \rotatebox[origin=c]{90}{traffic sign} & \rotatebox[origin=c]{90}{vegetation} & \rotatebox[origin=c]{90}{sky} & \rotatebox[origin=c]{90}{person} & \rotatebox[origin=c]{90}{rider} & \rotatebox[origin=c]{90}{car} & \rotatebox[origin=c]{90}{bus} & \rotatebox[origin=c]{90}{motorbike} & \rotatebox[origin=c]{90}{bicycle} & \rotatebox[origin=c]{90}{\bf mIoU} \\
				\midrule

				DCAN \cite{wu2018dcan} & \multirow{6}{*}{\rotatebox[origin=c]{90}{VGG16}} &
				79.9 & 30.4 & 70.8 & 1.6 & \bf 0.6 & 22.3 & 6.7 & 23.0 & 76.9 & 73.9 & 41.9 & 16.7 & 61.7 & 11.5 & \bf 10.3 & 38.6 & 35.4 \\

				DADA \cite{vu2019dada} &  &
				71.1 & 29.8 & 71.4 & 3.7 & 0.3 & \bf 33.2 & 6.4 & 15.6 & 81.2 & 78.9 & 52.7 & 13.1 &  75.9 & 25.5 & 10.0 & 20.5 & 36.8 \\

				GIO-Ada \cite{chen2019learning} &  &
				78.3 & 29.2 & 76.9 & \bf 11.4 & 0.3 & 26.5 & 10.8 & 17.2 & 81.7 & \bf 81.9 & 45.8 &
				15.4 & 68.0 & 15.9 & 7.5 & 30.4 & 37.3 \\

				TGCF-DA \cite{Choi2019self} &  &
				\bf 90.1 & \bf 48.6 & \bf 80.7 & 2.2 & 0.2 & 27.2 & 3.2 & 14.3 & \bf 82.1 & 78.4 & 54.4 & 16.4 & \bf 82.5 & 12.3 & 1.7 & 21.8 & 38.5 \\

				BDL \cite{li2019bidirectional} &  &
				72.0 & 30.3 & 74.5 & 0.1 & 0.3 & 24.6 & 10.2 & 25.2 & 80.5 & 80.0 &
				54.7 & 23.2 & 72.7 & 24.0 & 7.5 & 44.9 & 39.0 \\

				Ours &  &
				73.0 & 31.1 & 77.1 & 0.2 & 0.5 & 27.0 & \bf 11.3 & \bf 27.4 & 81.2 & 81.0 &
				\bf 59.0 & \bf 25.6 & 75.0 & \bf 26.3 & 10.1 & \bf 47.4 & \bf 40.8 \\

				\midrule

				SIBAN \cite{luo2019significance} & \multirow{6}{*}{\rotatebox[origin=c]{90}{ResNet101}} &
				82.5 & 24.0 & 79.4 & \xmark & \xmark & \xmark & 16.5 & 12.7 & 79.2 & 82.8 & 58.3 & 18.0 & 79.3 & 25.3 &
				17.6 & 25.9 & 46.3 \\

				CLAN \cite{luo2019taking} &  &
				81.3 & 37.0 & 80.1 & \xmark & \xmark & \xmark & 16.1 & 13.7 & 78.2 & 81.5 & 53.4 & 21.2 & 73.0 & 32.9 &
				22.6 & 30.7 & 47.8 \\

				MaxSquare \cite{maxsquareloss} &  &
				82.9 & 40.7 & 80.3 & \xmark & \xmark & \xmark & 12.8 & \bf 18.2 & \bf 82.5 & 82.2 & 53.1 & 18.0 & 79.0 & 31.4 & 10.4 & 35.6 & 48.2 \\

				DADA \cite{vu2019dada} &  &
				\bf 89.2 & 44.8 & \bf 81.4 & \xmark & \xmark & \xmark & 8.6 & 11.1 & 81.8 & \bf 84.0 & 54.7 & 19.3 & \bf 79.7 & 40.7 & 14.0 & 38.8 & 49.8 \\

				BDL \cite{li2019bidirectional} &  &
				86.0 & \bf 46.7 & 80.3 & \xmark & \xmark & \xmark & 14.1 & 11.6 & 79.2 & 81.3 & 54.1 & 27.9 & 73.7 & \bf 42.2 & 25.7 & \bf 45.3 & 51.4 \\

				Ours &  &
				82.5 & 42.2 & 81.3 & \xmark & \xmark & \xmark & \bf 18.3 & 15.9 & 80.6 & 83.5 &
				\bf 61.4 & \bf 33.2 & 72.9 & 39.3 & \bf 26.6 & 43.9 & \bf 52.4 \\

				\bottomrule
			\end{tabular}
		\end{center}
		\vspace{-0.3in}
	\end{table*}

	\subsection{Cross-Domain Spatial Attention Module}

	The goal of CD-SAM is to adapt spatial contextual information across two domains. To achieve this, we introduce the forward direction ("source-to-target") to augment source features by selectively aggregating target features based on their similarities. We further introduce the backward direction ("target-to-source") to update target features by aggregating source features in the same way.

	The architecture of CD-SAM is illustrated in Figure~\ref{fig:spatial_attention}. Given $ \textbf{A}_{s}\in \mathbb{R}^{C \times H \times W} $ and $ \textbf{A}_{t} \in \mathbb{R}^{C \times H \times W} $ ($ C $ denotes the channel number and $ H \times W $ indicates the spatial size), two parallel convolution layers are applied to generate $ \textbf{Q} \in \mathbb{R}^{C \times H \times W} $ and $ \textbf{K} \in \mathbb{R}^{C \times H \times W} $, respectively. $ \textbf{A}_{s} $ and $ \textbf{A}_{t} $ are also fed into another convolution layer to obtain $ \textbf{V}_s \in \mathbb{R}^{C \times H \times W} $ and $ \textbf{V}_t \in \mathbb{R}^{C \times H \times W} $. We reshape $ \textbf{Q} $, $ \textbf{V}_s $, $ \textbf{K} $, and $ \textbf{V}_t $ to $ C \times N $, where $ N = H \times W $. To determine spatial context relationships between each position in $ \textbf{A}_{s} $ and $ \textbf{A}_{t} $, an energy map $ \Phi \in \mathbb{R}^{ N \times N} $ is formulated as $ \Phi = \textbf{Q}^T \textbf{K} $, where $ \Phi^{(i, j)} $ measure the similarity between $ i^{th} $ position in $ \textbf{A}_{s} $ and $ j^{th} $ position in $ \textbf{A}_{t} $. To augment $ \textbf{A}_{s} $ with spatial context information from $ \textbf{A}_{t} $ and vice versa, a bidirectional feature adaptation is defined as follows.

	During the forward direction, we first define the "source-to-target" spatial attention map as,
	\begin{equation}
	\Gamma_{s{\rightarrow}t}^{(i, j)} = \frac{exp(\Phi^{(i, j)})}{\sum_{j=1}^{N_t} exp(\Phi^{(i, j)})},
	\end{equation}
	where $ \Gamma_{s{\rightarrow}t}^{(i, j)} $ indicates the impact of $ i^{th} $ position in $ \textbf{A}_{s} $ to $ j^{th} $ position in $ \textbf{A}_{t} $. To capture spatial context in the target domain, we update $ \textbf{A}_{s} $ as,
	\begin{equation}
	\textbf{A}_{s}^{'} = \textbf{A}_{s} + \lambda_{s} \textbf{V}_t \Gamma_{s{\rightarrow}t}^T,
	\end{equation}
	where $ \lambda_{s} $ leverages the importance of target-domain context and original source features. In this regime, each position in $ \textbf{A}_{s}^{'} $ has a global context view of target features.

	For the backward direction, the "target-to-source" spatial attention map is formulated as,
	\begin{equation}
	\Gamma_{t{\rightarrow}s}^{(i, j)} = \frac{exp(\Phi^{(i, j)})}{\sum_{i=1}^{N_s} exp(\Phi^{(i, j)})},
	\end{equation}
	where $ \Gamma_{t{\rightarrow}s}^{(i, j)} $ indicates to what extent the $ j^{th} $ position in $ \textbf{A}_{t} $ attends to the $ i^{th} $ position in $ \textbf{A}_{s} $. Similarly, $ \textbf{A}_{t} $ is updated by,
	\begin{equation}
	\textbf{A}_{t}^{'} = \textbf{A}_{t} + \lambda_{t} \textbf{V}_s \Gamma_{t{\rightarrow}s},
	\end{equation}
	where $ \lambda_{t} $ leverages the importance of source-domain context and original target features. As a consequence, each position in $ \textbf{A}_{s}^{'} $ and $ \textbf{A}_{t}^{'} $ is a combination of their original feature and the weighed sum of features from the opposite domain. Therefore, $ \textbf{A}_{s}^{'} $ and $ \textbf{A}_{t}^{'} $ allow us to encode the spatial context of both source and target domains.

	\subsection{Cross-Domain Channel Attention Module}

	Given $ \textbf{B}_{s}\in \mathbb{R}^{C \times H \times W} $ and $ \textbf{B}_{t} \in \mathbb{R}^{C \times H \times W} $, the CD-CAM is designed to adapt semantic context between source and target domains (Figure~\ref{fig:channel_attention}) by following the same bidirectional structure as CD-SAM. Different from CD-SAM that applies convolution layers to obtain $ \textbf{Q} $, $ \textbf{K} $, $ \textbf{V}_s $, and $ \textbf{V}_t $ before measuring spatial relationships. Here, $ \textbf{B}_{s} $ and $ \textbf{B}_{t} $ are directly used to capture their semantical context relationships, which allows us to maintain interdependencies between channel maps \cite{fu2019dual}. Specifically, we reshape both $ \textbf{B}_{s} $ and $ \textbf{B}_{t} $ to $ C \times N $, where $ N = H \times W $. The energy map is defined as $ \Theta = \textbf{B}_{t} \textbf{B}_{s}^{T} \in \mathbb{R}^{C \times C} $, where $ \Theta^{(i, j)} $ denotes the similarity between $ i^{th} $ channel in $ \textbf{B}_{s} $ and $ j^{th} $ channel in $ \textbf{B}_{t} $.

	For the forward direction, the "source-to-target" attention map is given by,
	\begin{equation}
	\Psi_{s{\rightarrow}t}^{(i, j)} = \frac{exp(\Theta^{(i, j)})}{\sum_{j=1}^{C} exp(\Theta^{(i, j)})},
	\end{equation}
	where $ \Psi_{s{\rightarrow}t}^{(i, j)} $ measures the impact of $ i^{th} $ channel in $ \textbf{B}_{s} $ to $ j^{th} $ channel in $ \textbf{B}_{t} $. To model the cross-domain semantic context dependencies, $ \textbf{B}_{s} $ is updated by,
	\begin{equation}
	\textbf{B}_{s}^{'} = \textbf{B}_{s} + \xi_{s} \Psi_{s{\rightarrow}t} \textbf{B}_{t},
	\end{equation}
	where $ \xi_{s} $ leverages the associations between target-domain semantic information and original source features. As a consequence, each channel in $ \textbf{B}_{s}^{'} $ is augmented by selectively aggregating semantic information from $ \textbf{B}_{t} $.

	During the backward direction, the "target-to-source" attention map is,
	\begin{equation}
	\Psi_{t{\rightarrow}s}^{(i, j)} = \frac{exp(\Theta^{(i, j)})}{\sum_{i=1}^{C} exp(\Theta^{(i, j)})}
	\end{equation}
	To take semantic context in $ \textbf{B}_{s} $ into consideration, we have
	\begin{equation}
	\textbf{B}_{t}^{'} = \textbf{B}_{t} + \xi_{t} \Psi_{t{\rightarrow}s}^T \textbf{B}_{s},
	\end{equation}
	where $ \xi_{t} $ leverages the associations between original target features and semantic contexts from the source domain. It is noteworthy that by considering cross-domain semantic context, our framework is able to further reduce domain discrepancy from the context perspective.

	\subsection{Aggregation of Spatial and Channel Context}

	To take full advantage of spatial and channel context information, we aggregate the outputs from these two cross-domain attention modules. Specifically, $ \textbf{A}_{s}^{'} $ and $ \textbf{B}_{s}^{'} $ are concatenated and then fed into a convolution layer to generate the enhanced source feature $ \textbf{Z}_{s} \in \mathbb{R}^{C \times H \times W} $. Obviously, $ \textbf{Z}_{s} $ is enriched by spatial and semantic context dependencies from both source and target domains. The same operation is also performed on $ \textbf{A}_{t}^{'} $ and $ \textbf{B}_{t}^{'} $ to obtain $ \textbf{Z}_{t} \in \mathbb{R}^{C \times H \times W} $.

	\newcommand{\cmark}{\ding{51}}%
	\begin{table}
		\caption{Ablation study on "GTA5 to Cityscapes".}
		\label{table:ablation_gta2city}

		\footnotesize
		\setlength\tabcolsep{13pt}
		\begin{center}
			\begin{tabularx}{.45\textwidth}{ c|c|c|c @{} }
				\toprule
				\multicolumn{4}{ c }{\bf GTA5 to Cityscapes } \\
				\midrule
				Base & CD-SAM & CD-CAM & \bf mIoU \\
				\midrule
				\multirow{4}{*}{VGG16}
				&  &  & 41.3 \\
				& \cmark &  & 43.7 \\
				&  &  \cmark & 43.6 \\
				& \cmark & \cmark & 44.9 \\

				\midrule
				\multirow{4}{*}{ResNet101}
				&  &  & 48.5 \\
				& \cmark &  & 49.0 \\
				&  &  \cmark & 48.8 \\
				& \cmark & \cmark & 49.2 \\
				\bottomrule
			\end{tabularx}
		\end{center}
		\vspace{-0.2in}
	\end{table}

	\subsection{Training Objective}

	Our framework contains a segmentation loss $ \mathcal{L}_{seg} $ and an adversarial loss $ \mathcal{L}_{adv} $. We first feed $ \textbf{Z}_s $ and $ \textbf{Z}_t $ into the classifier $ G $ to predict their segmentation outputs $ G(\textbf{Z}_s) $ and $ G(\textbf{Z}_t) $. The segmentation loss of $ G(\textbf{Z}_s) $ is defined as:
	\begin{equation}
	\mathcal{L}_{seg}(G(\textbf{Z}_s), Y_s) = -\sum_{i=1}^{H \times W}\sum_{j=1}^{L}Y_s^{(i, j)}G(\textbf{Z}_s)^{(i, j)},
	\end{equation}
	where $ L $ is the number of label classes. $ \mathcal{L}_{seg}(G(\textbf{Z}_t), Y_s^{st}) $ is defined in a similar way. To adapt structured output space \cite{tsai2018learning}, a discriminator $ D_1 $ is applied to $ G(\textbf{Z}_s) $ and $ G(\textbf{Z}_t) $ to make them be indistinguishable from each other. To achieve this, an adversarial loss $ \mathcal{L}_{adv}(G(\textbf{Z}_s), G(\textbf{Z}_t)) $ is formulated as,
	\begin{equation}
	\begin{aligned}
	\mathcal{L}_{adv}(G(\textbf{Z}_s), G(\textbf{Z}_t), D_1) = {} &
	\mathop{\mathbb{E}}[log{D_1(G(\textbf{Z}_s))}] + \\ &
	\mathop{\mathbb{E}}[log{(1 - D_1(G(\textbf{Z}_t)))}]
	\end{aligned}
	\end{equation}
	To encourage $ \textbf{A}_s^{'} $, $ \textbf{A}_t^{'} $, $ \textbf{B}_s^{'} $ and $ \textbf{B}_t^{'} $ to encode useful information for semantic segmentation, they are also fed into the classifier $ G $ to predict their segmentation outputs. The overall segmentation loss is given by,
	\begin{equation}
	\begin{aligned}
	\mathcal{L}_{seg}  = {} &
	\mathcal{L}_{seg}(G(\textbf{Z}_s), Y_s) + \mathcal{L}_{seg}(G(\textbf{Z}_t), Y_t^{st}) + \\ & \mathcal{L}_{seg}(G(\textbf{A}_s^{'}), Y_s) +  \mathcal{L}_{seg}(G(\textbf{A}_t^{'}), Y_t^{st}) + \\ & \mathcal{L}_{seg}(G(\textbf{B}_s^{'}), Y_s) + \mathcal{L}_{seg}(G(\textbf{B}_t^{'}), Y_t^{st})
	\end{aligned}
	\end{equation}
	We also encourage $ G(\textbf{A}_s^{'}) $ and $ G(\textbf{A}_t^{'}) $ to have similar structured layout, and enforce $ G(\textbf{B}_s^{'}) $ to be indistinguishable from $ G(\textbf{B}_t^{'}) $. Therefore, the overall adversarial loss can be written as,
	\begin{equation}
	\begin{aligned}
	\mathcal{L}_{adv} = {} &
	\mathcal{L}_{adv}(G(\textbf{Z}_s), G(\textbf{Z}_t), D_1) + \\ &
	\mathcal{L}_{adv}(G(\textbf{A}_s^{'}), G(\textbf{A}_t^{'}), D_2) + \\ &
	\mathcal{L}_{adv}(G(\textbf{B}_s^{'}), G(\textbf{B}_t^{'}), D_3),
	\end{aligned}
	\end{equation}
	where $ D_2 $ and $ D_3 $ are two discriminators. Specifically, $ D_2 $ aims to discriminate between $ G(\textbf{A}_s^{'})$ and $ G(\textbf{A}_t^{'}) $, while $ D_3 $ attempts to distinguish between $ G(\textbf{B}_s^{'})$ and $ G(\textbf{B}_t^{'}) $.

	Taken them together, the training objective of our framework is:
	\begin{equation}
	\min\limits_{E, G} \max\limits_{D_1, D_2, D_3} \mathcal{L}_{seg} + \lambda \mathcal{L}_{adv}
	\end{equation}
	where $ \lambda $ controls the importance of $ \mathcal{L}_{seg} $ and $ \mathcal{L}_{adv} $.

	\begin{table}
		\caption{Ablation study on "SYNTHIA to Cityscapes".}
		\label{table:ablation_synthia2city}

		\footnotesize
		\setlength\tabcolsep{13pt}
		\begin{center}
			\begin{tabularx}{.45\textwidth}{ c|c|c|c @{} }
				\toprule
				\multicolumn{4}{ c }{\bf SYNTHIA to Cityscapes } \\
				\midrule
				Base & CD-SAM & CD-CAM & \bf mIoU \\
				\midrule
				\multirow{4}{*}{VGG16}
				&  &  & 39.0 \\
				& \cmark &  & 40.2 \\
				&  &  \cmark & 40.0 \\
				& \cmark & \cmark & 40.8 \\

				\midrule
				\multirow{4}{*}{ResNet101}
				&  &  & 51.4 \\
				& \cmark &  & 51.8 \\
				&  &  \cmark & 52.0 \\
				& \cmark & \cmark & 52.4 \\
				\bottomrule
			\end{tabularx}
		\end{center}
		\vspace{-0.2in}
	\end{table}

	\section{Experiments}

	In this section, we evaluate our method on synthetic-to-real domain adaptation for urban scene understanding problem. Extensive empirical experiments and ablation studies are performed to demonstrate out method's superiority over existing state-of-the-art models.
	We also visualize the cross-domain attention maps to reveal context dependencies between source and target domains.

	\subsection{Datasets}

	Two synthetic datasets, \ie, GTA5 \cite{richter2016playing} and SYNTHIA-RAND-CITYSCAPES \cite{ros2016synthia} are used as the source domain in our study, while the Cityscapes \cite{cordts2016cityscapes} is served as the target domain. Specifically, the GTA5 is collected from a photorealistic open-world game known as Grand Theft Auto V, which contains 24,966 images with pixel-accurate semantic labels. The resolution of each image is 1914 $ \times $ 1052. SYNTHIA-RAND-CITYSCAPES contains 9,400 images (1280 $ \times $ 760) with precise pixel-level semantic annotations, which are generated from a virtual city. Cityscapes is a large-scale street scene datasets collected from 50 cities, including 5,000 images with high-quality pixel-level annotations. These images are split into training (2,975 images), validation (500 images), and test (1,525 images) set, each of which with the resolution of 2048 $ \times $ 1024. Following the same setting as previous studies, only the training set from Cityscapes is used as the target domain, and the validation set is used for performance evaluation.

	\begin{figure*}
		\begin{center}
			\includegraphics[width=1.0\linewidth]{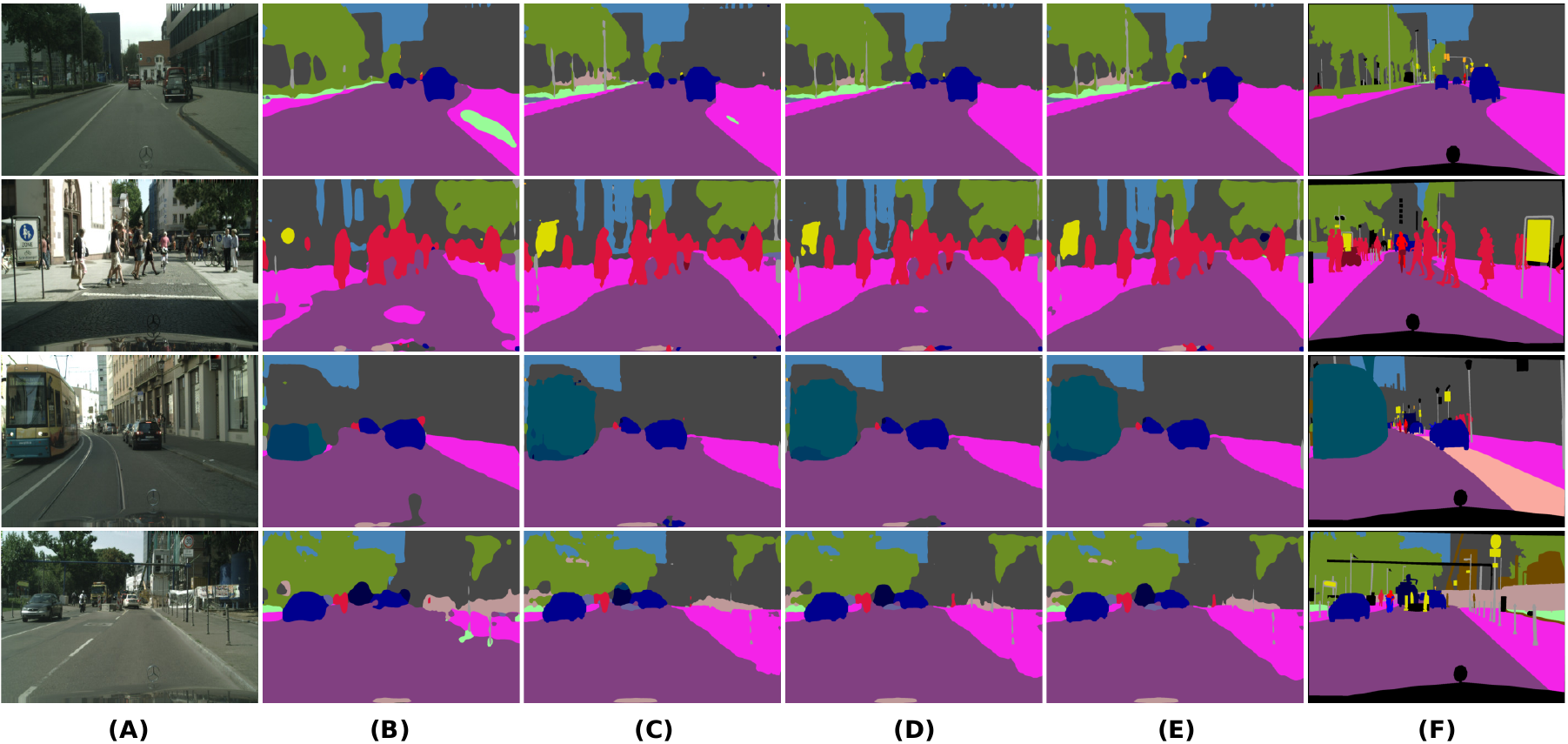}
		\end{center}
		\caption{Qualitative comparison between our method and the baseline model BDL \cite{li2019bidirectional}. For each given image (A), we present its segmentation output from (B) BDL, (C) our method incorporating CD-SAM only, (D) our method incorporating CD-CAM only, (E) our method considering both CD-SAM and CD-CAM, and the ground truth (F).}
		\label{fig:quality}
		\vspace{-0.2in}
	\end{figure*}

	\subsection{Implementation Details}
	\paragraph{Network Architecture}

	The same CycleGAN architecture \cite{zhu2017unpaired} as reported in BDL \cite{li2019bidirectional} is used to translate images from the source domain to the target domain. DeepLab-VGG16 and DeepLab-ResNet101, which are pre-trained on ImageNet \cite{deng2009imagenet}, are used as our segmentation network by following the same setting in \cite{tsai2018learning}. Both of them use DeepLab-v2 \cite{chen2018deeplab} as classifier, while DeepLab-VGG16 uses VGG16 \cite{simonyan2014very} and DeepLab-ResNet101 uses ResNet101 \cite{he2016deep} as the feature extractor. The three discriminators used for structured output adaptation have the identical architecture, each of which has 5 convolution layers with kernel 4$ \times $4 and stride of 2. The channel number of each layer is \{64, 128, 256, 512, 1\}. Each layer is followed by a leaky ReLU \cite{maas2013rectifier} parameterized by 0.2 except the last one. The CD-SAM contains 3 convolution layers with kernel 1$ \times $1 and stride of 1 to obtain the query and key-value pairs. The channel number of these convolution layers are \{128, 128, 1024\} and \{256, 256, 2048\} for DeepLab-VGG16 and DeepLab-ResNet101, respectively.

	\paragraph{Network Training}

	To train the CycleGAN network, we follow the same setting in BDL \cite{li2019bidirectional}. DeepLab-VGG16 is trained using Adam optimizer with initial learning rate 1e-5 and momentum (0.9, 0.99). We apply step decay to the learning rate with step size 50000 and drop factor 0.1. Both DeepLab-ResNet101 and CD-SAM use Stochastic Gradient Descent (SGD) optimizer with momentum 0.9 and weight decay 5e-4. The initial learning rate for DeepLab-ResNet101 and CD-SAM are 2.5e-4 and 1e-4, respectively, and are decreased by the same polynomial policy with power 0.9. For the discriminator, we use an Adam optimizer with momentum (0.9, 0.99). Its initial learning rate is set to 1e-6 for DeepLab-VGG16 and 1e-4 for DeepLab-ResNet101, respectively.


	\subsection{Performance Comparison}
	\paragraph{GTA5 to Cityscapes}

	Our method is first evaluated by using GTA5 as the source domain and Cityscapes as the target domain. The performance is assessed on 19 common classes between these two datasets by following the same evaluation criterion in previous studies \cite{li2019bidirectional, chen2019crdoco}. Our method is compared with existing state-of-the-art models by using VGG16 and ResNet101 as the base architectures. As shown in Table \ref{table:gta2city}, our method achieves the best performance compared to other models. Specifically, we surpass the mean intersection-over-union (mIoU) of feature alignment-based \cite{hoffman2016fcns, sankaranarayanan2018learning, luo2019taking} and curriculum-based methods \cite{zhang2017curriculum} by a large margin. This observation indicates that simply aligning feature space and label distribution cannot fully transfer domain knowledge in semantic segmentation. Compared to the models \cite{hoffman2017cycada, chen2019crdoco, li2019bidirectional} that are based on image-to-image translation, our method gains up to 9.5\% improvement by using VGG16, revealing that domain discrepancy can be further reduced by considering context adaptation. Similar to \cite{tsai2018learning, li2019bidirectional}, we also adapt structured output space in our model, but our method achieves significant performance improvement. This observation reveals the important role of context adaptation in knowledge transfer. It is noteworthy that the prediction of the "train" class is extremely challenging, owing to the limited "train" samples in the source domain. Our method enables to alleviate this limitation by adapting cross-domain context information. Compared to the CyCADA \cite{hoffman2017cycada}, we achieve 16.1\% improvement on the "train" class.

	\begin{figure}[t]
		\begin{center}
			\includegraphics[width=1.0\linewidth]{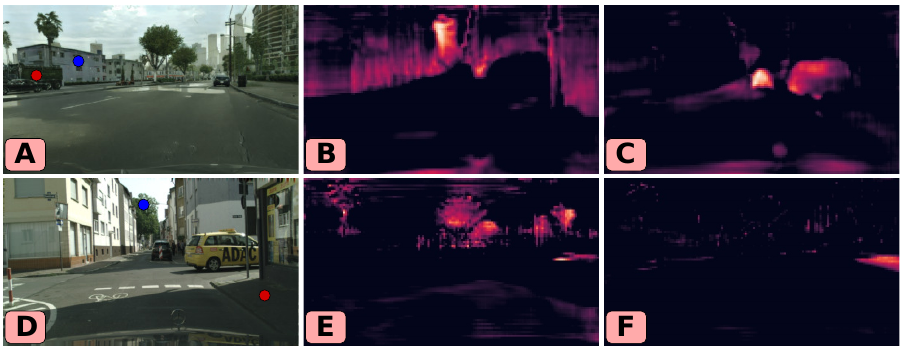}
		\end{center}
		\caption{An example of the spatial attention map. Given a source image (A) and a target image (D), we present the source-to-target attention maps (B) and (C) for the blue and red point in (A), respectively. Similarly, we present the target-to-source attention maps (E) and (F) of the blue and red point in (D), respectively. }
		\label{fig:spatial_attention_map}
		\vspace{-0.2in}
	\end{figure}

	\paragraph{SYNTHIA to Cityscapes}

	The superiority of our method is further proved on "SYNTHIA to Cityscapes". It is noteworthy that domain adaptation on "SYNTHIA to Cityscapes" is more challenging than "GTA5 to Cityscapes", owing to the large domain gap between these two domains. Following \cite{li2019bidirectional}, we consider the 16 and 13 common classes for VGG16 and ResNet101-based models, respectively. As summarized in Table \ref{table:synthia2city}, our method substantially outperforms other competitive models. Notably, we achieve a performance improvement of 1.8\% and 1.0\% over BDL \cite{li2019bidirectional} with VGG16 and ResNet101 base architectures. This result demonstrates the benefit of explicitly adapting cross-domain context dependencies in semantic segmentation, especially for two domains with significant differences.

	\subsection{Ablation Study}

	In this section, we conduct extensive ablation studies to investigate the effectiveness of two cross-domain attention modules in our model.

	\paragraph{GTA5 to Cityscapes}

	By incorporating CD-SAM and CD-CAM individually, we get 2.4\% and 2.3\% performance boost over the VGG16-based baseline (Table \ref{table:ablation_gta2city}). Taken them together, the mIoU is further improved to 44.9 mIoU. Similarly, 0.5\% and 0.3\% improvement is also observed in the ResNet101-based model by considering CD-SAM and CD-CAM. We achieve 49.2 mIoU by integrating both attention modules. To qualitatively demonstrate the superiority of our method, we showcase the examples of its segmentation outputs at different stages in Figure~\ref{fig:quality}. As shown in the figure, our method enables to predict more consistent segmentation outputs than the baseline model and becomes increasingly accurate by incorporating two cross-domain attention modules.

	\paragraph{SYNTHIA to Cityscapes}

	For VGG16-based model, CD-SAM and CD-CAM contribute to 1.2\% and 1.0\% improvement compared to the baseline (Table \ref{table:ablation_synthia2city}). Our method gains 1.8\% improvement by combining them. By applying CD-SAM and CD-CAM to ResNet101, we achieve 51.8 and 52.0 mIoU with 0.4\% and 0.6\% improvement over the baseline, respectively. It is further boosted to 52.4 mIoU when both of them are considered.

	Our results reveal that the proposed cross-attention mechanism significantly contributes to domain adaptation in semantic segmentation by adapting context dependencies. Furthermore, the two cross-domain attention modules play a complementary role in capturing context information.

	\subsection{Visualization of the Cross Attention}

	To fully understand the cross-attention mechanism in our model, we visualize the spatial attention maps in this section. As shown in Figure~\ref{fig:spatial_attention_map}, two images are randomly selected from the source and target domain. Recall that each position in the source feature has a spatial attention map corresponding to all positions in the target feature, and vice versa. We, therefore, select two positions in the source image and visualize their "source-to-target" attention map. For the blue point that is marked on a building in the source image, its spatial attention map mainly corresponds to the building in the target image. Similarly, we select another two positions in the target image and conduct the visualization of the "target-to-source" attention map. For the blue point in the target image, its attention map focuses on the vegetation in the source image. These visualizations demonstrate the power of our method in capturing cross-domain spatial context information.

	\begin{table}
		\caption{Ablation study of $ \lambda_{s} $, $ \lambda_{t} $, $ \xi_{s} $, and $ \xi_{t} $.}
		\label{table:ablation_attention}

		\footnotesize
		\setlength\tabcolsep{17pt}
		\begin{center}
			\begin{tabularx}{.45\textwidth}{ cccc @{} }
				\toprule
				$ \lambda_{s}/\lambda_{t}/\xi_{s}/\xi_{t} $ & 0.1 & 1 & 10 \\
				\midrule
				mIoU & 43.7 & 44.9 & 40.6 \\
				\bottomrule
			\end{tabularx}
		\end{center}
		\vspace{-0.2in}
	\end{table}

	\subsection{Parameter Sensitivity Analysis}

	In this section, we perform a sensitivity analysis of $ \lambda_{s} $, $ \lambda_{t} $, $ \xi_{s} $, and $ \xi_{t} $ as shown in Table \ref{table:ablation_attention}. We investigate three different choices, \ie, 0.1, 1, and 10, indicating how much attention should pay for the context information from the opposite domain. Our results reveal that $ \lambda_{s}=\lambda_{t}=\xi_{s}=\xi_{t}=1 $ performs best. The reason is that a small value fails to capture cross-domain context dependencies, while a large value may disturb the original feature.

	\section{Conclusion}

	In this paper, we propose an innovative cross-attention mechanism for domain adaptation
	by adapting the semantic context. Specifically, we introduce two cross-domain attention modules to capture spatial and channel context between source and target domains. The obtained contextual dependencies, which are shared across two domains, are further adapted to decrease the domain discrepancy. Empirical studies demonstrate that our method achieves the new state-of-the-art performance on "GTA5-to-Cityscapes" and "SYNTHIA-to-Cityscapes".

	{\small
		\bibliographystyle{ieee_fullname}
		\bibliography{egbib}

\begin{thebibliography}{10}\itemsep=-1pt

\bibitem{chang2019all}
Wei-Lun Chang, Hui-Po Wang, Wen-Hsiao Peng, and Wei-Chen Chiu.
\newblock All about structure: Adapting structural information across domains
  for boosting semantic segmentation.
\newblock In {\em Proceedings of the IEEE Conference on Computer Vision and
  Pattern Recognition}, pages 1900--1909, 2019.

\bibitem{chen2018deeplab}
Liang-Chieh Chen, George Papandreou, Iasonas Kokkinos, Kevin Murphy, and Alan~L
  Yuille.
\newblock Deeplab: Semantic image segmentation with deep convolutional nets,
  atrous convolution, and fully connected crfs.
\newblock {\em IEEE transactions on pattern analysis and machine intelligence},
  40(4):834--848, 2018.

\bibitem{chen2019learning}
Yuhua Chen, Wen Li, Xiaoran Chen, and Luc~Van Gool.
\newblock Learning semantic segmentation from synthetic data: A geometrically
  guided input-output adaptation approach.
\newblock In {\em Proceedings of the IEEE Conference on Computer Vision and
  Pattern Recognition}, pages 1841--1850, 2019.

\bibitem{chen2019crdoco}
Yun-Chun Chen, Yen-Yu Lin, Ming-Hsuan Yang, and Jia-Bin Huang.
\newblock Crdoco: Pixel-level domain transfer with cross-domain consistency.
\newblock In {\em Proceedings of the IEEE Conference on Computer Vision and
  Pattern Recognition}, pages 1791--1800, 2019.

\bibitem{Choi2019self}
Jaehoon Choi, Taekyung Kim, and Changick Kim.
\newblock Self-ensembling with gan-based data augmentation for domain
  adaptation in semantic segmentation.
\newblock {\em arXiv preprint arXiv:1909.00589}, 2019.

\bibitem{cordts2016cityscapes}
Marius Cordts, Mohamed Omran, Sebastian Ramos, Timo Rehfeld, Markus Enzweiler,
  Rodrigo Benenson, Uwe Franke, Stefan Roth, and Bernt Schiele.
\newblock The cityscapes dataset for semantic urban scene understanding.
\newblock In {\em Proceedings of the IEEE conference on computer vision and
  pattern recognition}, pages 3213--3223, 2016.

\bibitem{deng2009imagenet}
Jia Deng, Wei Dong, Richard Socher, Li-Jia Li, Kai Li, and Li Fei-Fei.
\newblock Imagenet: A large-scale hierarchical image database.
\newblock In {\em 2009 IEEE conference on computer vision and pattern
  recognition}, pages 248--255. Ieee, 2009.

\bibitem{fu2019dual}
Jun Fu, Jing Liu, Haijie Tian, Yong Li, Yongjun Bao, Zhiwei Fang, and Hanqing
  Lu.
\newblock Dual attention network for scene segmentation.
\newblock In {\em Proceedings of the IEEE Conference on Computer Vision and
  Pattern Recognition}, pages 3146--3154, 2019.

\bibitem{ganin2014unsupervised}
Yaroslav Ganin and Victor Lempitsky.
\newblock Unsupervised domain adaptation by backpropagation.
\newblock {\em arXiv preprint arXiv:1409.7495}, 2014.

\bibitem{goodfellow2014generative}
Ian Goodfellow, Jean Pouget-Abadie, Mehdi Mirza, Bing Xu, David Warde-Farley,
  Sherjil Ozair, Aaron Courville, and Yoshua Bengio.
\newblock Generative adversarial nets.
\newblock In {\em Advances in neural information processing systems}, pages
  2672--2680, 2014.

\bibitem{he2016deep}
Kaiming He, Xiangyu Zhang, Shaoqing Ren, and Jian Sun.
\newblock Deep residual learning for image recognition.
\newblock In {\em Proceedings of the IEEE conference on computer vision and
  pattern recognition}, pages 770--778, 2016.

\bibitem{hoffman2017cycada}
Judy Hoffman, Eric Tzeng, Taesung Park, Jun-Yan Zhu, Phillip Isola, Kate
  Saenko, Alexei~A Efros, and Trevor Darrell.
\newblock Cycada: Cycle-consistent adversarial domain adaptation.
\newblock {\em arXiv preprint arXiv:1711.03213}, 2017.

\bibitem{hoffman2016fcns}
Judy Hoffman, Dequan Wang, Fisher Yu, and Trevor Darrell.
\newblock Fcns in the wild: Pixel-level adversarial and constraint-based
  adaptation.
\newblock {\em arXiv preprint arXiv:1612.02649}, 2016.

\bibitem{jang2019learning}
Yunhun Jang, Hankook Lee, Sung~Ju Hwang, and Jinwoo Shin.
\newblock Learning what and where to transfer.
\newblock {\em arXiv preprint arXiv:1905.05901}, 2019.

\bibitem{li2019bidirectional}
Yunsheng Li, Lu Yuan, and Nuno Vasconcelos.
\newblock Bidirectional learning for domain adaptation of semantic
  segmentation.
\newblock In {\em Proceedings of the IEEE Conference on Computer Vision and
  Pattern Recognition}, pages 6936--6945, 2019.

\bibitem{lian2019constructing}
Qing Lian, Fengmao Lv, Lixin Duan, and Boqing Gong.
\newblock Constructing self-motivated pyramid curriculums for cross-domain
  semantic segmentation: A non-adversarial approach.
\newblock {\em arXiv preprint arXiv:1908.09547}, 2019.

\bibitem{lin2018multi}
Di Lin, Yuanfeng Ji, Dani Lischinski, Daniel Cohen-Or, and Hui Huang.
\newblock Multi-scale context intertwining for semantic segmentation.
\newblock In {\em Proceedings of the European Conference on Computer Vision
  (ECCV)}, pages 603--619, 2018.

\bibitem{lin2017structured}
Zhouhan Lin, Minwei Feng, Cicero Nogueira~dos Santos, Mo Yu, Bing Xiang, Bowen
  Zhou, and Yoshua Bengio.
\newblock A structured self-attentive sentence embedding.
\newblock {\em arXiv preprint arXiv:1703.03130}, 2017.

\bibitem{long2015fully}
Jonathan Long, Evan Shelhamer, and Trevor Darrell.
\newblock Fully convolutional networks for semantic segmentation.
\newblock In {\em Proceedings of the IEEE conference on computer vision and
  pattern recognition}, pages 3431--3440, 2015.

\bibitem{long2015learning}
Mingsheng Long, Yue Cao, Jianmin Wang, and Michael~I Jordan.
\newblock Learning transferable features with deep adaptation networks.
\newblock {\em arXiv preprint arXiv:1502.02791}, 2015.

\bibitem{long2017deep}
Mingsheng Long, Han Zhu, Jianmin Wang, and Michael~I Jordan.
\newblock Deep transfer learning with joint adaptation networks.
\newblock In {\em Proceedings of the 34th International Conference on Machine
  Learning-Volume 70}, pages 2208--2217. JMLR. org, 2017.

\bibitem{luo2019significance}
Yawei Luo, Ping Liu, Tao Guan, Junqing Yu, and Yi Yang.
\newblock Significance-aware information bottleneck for domain adaptive
  semantic segmentation.
\newblock {\em arXiv preprint arXiv:1904.00876}, 2019.

\bibitem{luo2019taking}
Yawei Luo, Liang Zheng, Tao Guan, Junqing Yu, and Yi Yang.
\newblock Taking a closer look at domain shift: Category-level adversaries for
  semantics consistent domain adaptation.
\newblock In {\em Proceedings of the IEEE Conference on Computer Vision and
  Pattern Recognition}, pages 2507--2516, 2019.

\bibitem{maas2013rectifier}
Andrew~L Maas, Awni~Y Hannun, and Andrew~Y Ng.
\newblock Rectifier nonlinearities improve neural network acoustic models.
\newblock In {\em Proc. icml}, volume~30, page~3, 2013.

\bibitem{maxsquareloss}
Hongyang~Xue Minghao~Chen and Deng Cai.
\newblock Domain adaptation for semantic segmentation with maximum squares
  loss.
\newblock In {\em IEEE International Conference on Computer Vision(ICCV)},
  2019.

\bibitem{mottaghi2014role}
Roozbeh Mottaghi, Xianjie Chen, Xiaobai Liu, Nam-Gyu Cho, Seong-Whan Lee, Sanja
  Fidler, Raquel Urtasun, and Alan Yuille.
\newblock The role of context for object detection and semantic segmentation in
  the wild.
\newblock In {\em Proceedings of the IEEE Conference on Computer Vision and
  Pattern Recognition}, pages 891--898, 2014.

\bibitem{pan2009survey}
Sinno~Jialin Pan and Qiang Yang.
\newblock A survey on transfer learning.
\newblock {\em IEEE Transactions on knowledge and data engineering},
  22(10):1345--1359, 2009.

\bibitem{richter2016playing}
Stephan~R Richter, Vibhav Vineet, Stefan Roth, and Vladlen Koltun.
\newblock Playing for data: Ground truth from computer games.
\newblock In {\em European Conference on Computer Vision}, pages 102--118.
  Springer, 2016.

\bibitem{ros2016synthia}
German Ros, Laura Sellart, Joanna Materzynska, David Vazquez, and Antonio~M
  Lopez.
\newblock The synthia dataset: A large collection of synthetic images for
  semantic segmentation of urban scenes.
\newblock In {\em Proceedings of the IEEE conference on computer vision and
  pattern recognition}, pages 3234--3243, 2016.

\bibitem{sankaranarayanan2018learning}
Swami Sankaranarayanan, Yogesh Balaji, Arpit Jain, Ser Nam~Lim, and Rama
  Chellappa.
\newblock Learning from synthetic data: Addressing domain shift for semantic
  segmentation.
\newblock In {\em Proceedings of the IEEE Conference on Computer Vision and
  Pattern Recognition}, pages 3752--3761, 2018.

\bibitem{simonyan2014very}
Karen Simonyan and Andrew Zisserman.
\newblock Very deep convolutional networks for large-scale image recognition.
\newblock {\em arXiv preprint arXiv:1409.1556}, 2014.

\bibitem{tsai2018learning}
Yi-Hsuan Tsai, Wei-Chih Hung, Samuel Schulter, Kihyuk Sohn, Ming-Hsuan Yang,
  and Manmohan Chandraker.
\newblock Learning to adapt structured output space for semantic segmentation.
\newblock {\em arXiv preprint arXiv:1802.10349}, 2018.

\bibitem{tzeng2017adversarial}
Eric Tzeng, Judy Hoffman, Kate Saenko, and Trevor Darrell.
\newblock Adversarial discriminative domain adaptation.
\newblock In {\em Proceedings of the IEEE Conference on Computer Vision and
  Pattern Recognition}, pages 7167--7176, 2017.

\bibitem{tzeng2014deep}
Eric Tzeng, Judy Hoffman, Ning Zhang, Kate Saenko, and Trevor Darrell.
\newblock Deep domain confusion: Maximizing for domain invariance.
\newblock {\em arXiv preprint arXiv:1412.3474}, 2014.

\bibitem{vaswani2017attention}
Ashish Vaswani, Noam Shazeer, Niki Parmar, Jakob Uszkoreit, Llion Jones,
  Aidan~N Gomez, {\L}ukasz Kaiser, and Illia Polosukhin.
\newblock Attention is all you need.
\newblock In {\em Advances in neural information processing systems}, pages
  5998--6008, 2017.

\bibitem{vu2019dada}
Tuan-Hung Vu, Himalaya Jain, Maxime Bucher, Matthieu Cord, and Patrick
  P{\'e}rez.
\newblock Dada: Depth-aware domain adaptation in semantic segmentation.
\newblock {\em arXiv preprint arXiv:1904.01886}, 2019.

\bibitem{wu2018dcan}
Zuxuan Wu, Xintong Han, Yen-Liang Lin, Mustafa Gokhan~Uzunbas, Tom Goldstein,
  Ser Nam~Lim, and Larry~S Davis.
\newblock Dcan: Dual channel-wise alignment networks for unsupervised scene
  adaptation.
\newblock In {\em Proceedings of the European Conference on Computer Vision
  (ECCV)}, pages 518--534, 2018.

\bibitem{ying2018transfer}
Wei Ying, Yu Zhang, Junzhou Huang, and Qiang Yang.
\newblock Transfer learning via learning to transfer.
\newblock In {\em International Conference on Machine Learning}, pages
  5072--5081, 2018.

\bibitem{zhang2018context}
Hang Zhang, Kristin Dana, Jianping Shi, Zhongyue Zhang, Xiaogang Wang, Ambrish
  Tyagi, and Amit Agrawal.
\newblock Context encoding for semantic segmentation.
\newblock In {\em Proceedings of the IEEE Conference on Computer Vision and
  Pattern Recognition}, pages 7151--7160, 2018.

\bibitem{zhang2018self}
Han Zhang, Ian Goodfellow, Dimitris Metaxas, and Augustus Odena.
\newblock Self-attention generative adversarial networks.
\newblock {\em arXiv preprint arXiv:1805.08318}, 2018.

\bibitem{zhang2017curriculum}
Yang Zhang, Philip David, and Boqing Gong.
\newblock Curriculum domain adaptation for semantic segmentation of urban
  scenes.
\newblock In {\em The IEEE International Conference on Computer Vision (ICCV)},
  volume~2, page~6, 2017.

\bibitem{zhao2018psanet}
Hengshuang Zhao, Yi Zhang, Shu Liu, Jianping Shi, Chen Change~Loy, Dahua Lin,
  and Jiaya Jia.
\newblock Psanet: Point-wise spatial attention network for scene parsing.
\newblock In {\em Proceedings of the European Conference on Computer Vision
  (ECCV)}, pages 267--283, 2018.

\bibitem{zhu2017unpaired}
Jun-Yan Zhu, Taesung Park, Phillip Isola, and Alexei~A Efros.
\newblock Unpaired image-to-image translation using cycle-consistent
  adversarial networks.
\newblock In {\em Proceedings of the IEEE international conference on computer
  vision}, pages 2223--2232, 2017.

\bibitem{zou2018unsupervised}
Yang Zou, Zhiding Yu, BVK~Vijaya Kumar, and Jinsong Wang.
\newblock Unsupervised domain adaptation for semantic segmentation via
  class-balanced self-training.
\newblock In {\em European Conference on Computer Vision}, pages 297--313.
  Springer, 2018.

\end{thebibliography}
	}

\end{document}